\documentclass[10pt,twocolumn,letterpaper]{article}

\usepackage{iccv}
\usepackage{times}
\usepackage{graphicx}
\usepackage[caption=false]{subfig}
\usepackage{amsmath,amssymb}
\usepackage{algorithm,algorithmic}
\usepackage[noblocks]{authblk}

\usepackage[american]{babel}

\usepackage[pagebackref=true,breaklinks=true,colorlinks,bookmarks=false]{hyperref}

\iccvfinalcopy 

\ificcvfinal\fi
\begin{document}

\title{Person Re-identification with Correspondence Structure Learning}

\author[1]{\tt\small Yang Shen}
\author[1]{\tt\small Weiyao Lin}
\author[1]{\tt\small Junchi Yan}
\author[2]{\tt\small Mingliang Xu}
\author[3]{\tt\small Jianxin Wu}
\author[4]{\tt\small Jingdong Wang}
\affil[1]{\tt\small Shanghai Jiao Tong University, China\qquad${}^{2}$Zhengzhou University, China}
\affil[3]{\tt\small Nanjing University, China\qquad${}^{4}$Microsoft Research Asia}

\renewcommand\Authands{ and }

\date{}
\maketitle

\begin{abstract}
  This paper addresses the problem of handling spatial misalignments due to camera-view changes or human-pose variations in person re-identification. We first introduce a boosting-based approach to learn a correspondence structure which indicates the patch-wise matching probabilities  between images from a target camera pair. The learned correspondence structure can not only capture the spatial correspondence pattern between cameras but also handle the viewpoint or human-pose variation in individual images. We further introduce a global-based matching process. It integrates a global matching constraint over the learned correspondence structure to exclude cross-view misalignments during the image patch matching process, hence achieving a more reliable matching score between images. Experimental results on various datasets demonstrate the effectiveness of our approach.
\end{abstract}

\section{Introduction\label{section:introduction}}

\begin{figure}[t]
  \centering
  \subfloat[]{\includegraphics[width=3.0cm,height=2.8cm]{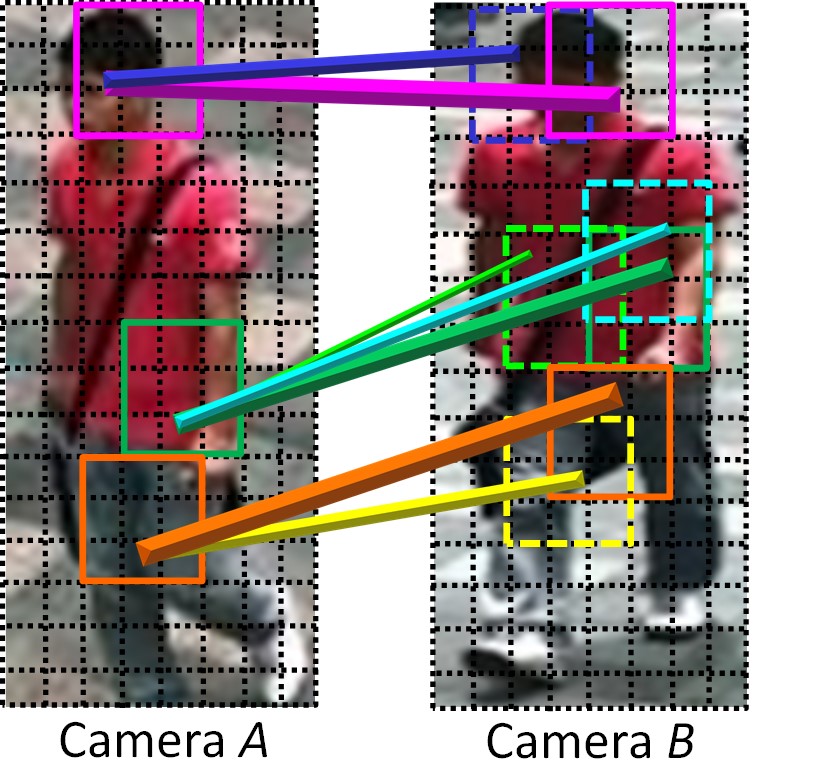}
    \label{fig:mapping structure_example_a}}
  \hspace{-3.2mm}
  \subfloat[]{\includegraphics[width=2.8cm,height=2.8cm]{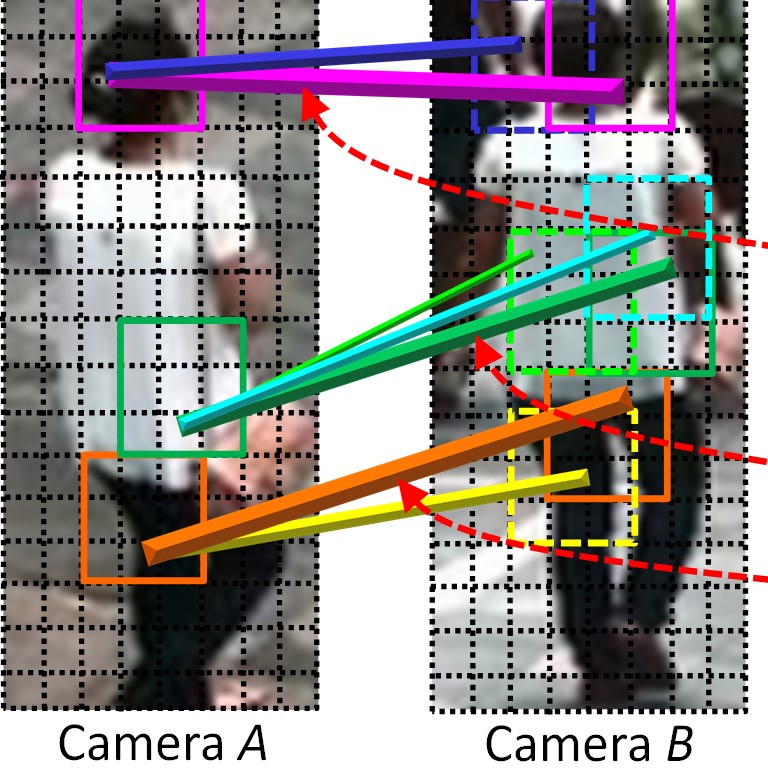}
    \label{fig:mapping structure_example_b}}
  \hspace{-1.8mm}
  \subfloat[]{\includegraphics[width=2.6cm,height=2.8cm]{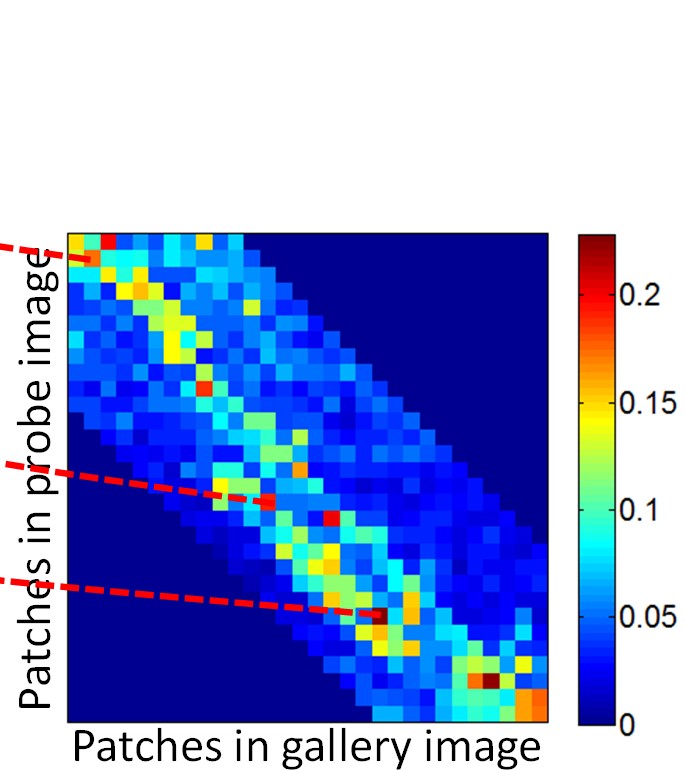}
    \label{fig:mapping structure_example_c}}
  \caption{(a) and (b): Two examples of using a correspondence structure to handle spatial misalignments between images from a camera pair. Images are obtained from the same camera pair: $A$ and $B$. The colored squares represent sample patches in each image while the lines between images indicate the matching probability between patches (line width is proportional to the probability values). (c): The correspondence structure matrix including all patch matching probabilities between $A$ and $B$ (the matrix is down-sampled for a clearer illustration). (Best viewed in color)
  }\label{fig:mapping structure_example}
\end{figure}

Person re-identification (Re-ID) is of increasing importance in visual surveillance. The goal of person Re-ID is to identify a specific person indicated by a probe image from a set of gallery images captured from cross-view cameras (i.e., cameras that are non-overlapping and different from the probe image's camera).\footnote{In this paper, an image refers to the pixel region of one person which is cropped from a larger image of a camera view (cf. Fig.~\ref{fig:mapping structure_example})~\cite{viper}.} It remains challenging due to the large appearance changes in different camera views and the interferences from background or object occlusion.

One major challenge for person Re-ID is the uncontrolled spatial misalignment between images due to camera-view changes or human-pose variations. For example, in Fig.~\ref{fig:mapping structure_example_a}, the green patch located in the lower part in camera $A$'s image corresponds to patches from the upper part in camera $B$'s image. However, most existing works \cite{SalientColor, rankboost, color_in,3,eiml,1,kernel-based,prid} focus on handling the overall appearance variations between images, while the spatial misalignment among images' local patches is not addressed. Although some patch-based methods \cite{part,12,6} address the spatial misalignment problem by decomposing images into patches and performing an online patch-level matching, their performances are often restrained by the online matching process which is easily affected by the mismatched patches due to similar appearance or occlusion.

In this paper, we argue that due to the stable setting of most cameras (e.g., fixed camera angle or location), each camera has a stable constraint on the spatial configuration of its captured images. For example, images in Figures~\ref{fig:mapping structure_example_a} and~\ref{fig:mapping structure_example_b} are obtained from the same camera pair: $A$ and $B$. Due to the constraint from camera angle difference, body parts in camera $A$'s images are located at lower places than those in camera $B$, implying a lower-to-upper correspondence pattern between them. Meanwhile, constraints from camera locations can also be observed. Camera $A$ (which monitors an exit region) includes more side-view images, while camera $B$ (monitoring a road) shows more front or back-view images. This further results in a high probability of side-to-front/back correspondence pattern.

Based on this intuition, we propose to learn a correspondence structure (i.e., a matrix including \emph{all} patch-wise matching probabilities between a camera pair, as Fig.~\ref{fig:mapping structure_example_c}) to encode the spatial correspondence pattern constrained by a camera pair, and utilize it to guide the patch matching and matching score calculation processes between images. With this correspondence structure, spatial misalignments can be suitably handled and patch matching results are less interfered by the confusion from appearance or occlusion. In order for the correspondence structure to model human-pose variations or local viewpoint changes inside a camera view, the correspondence structure for each patch is described by a one-to-many graph whose weights indicate the matching probabilities between patches, as in Fig.~\ref{fig:mapping structure_example}. Besides, a global constraint is also integrated during the patch matching process, so as to achieve a more reliable matching score between images. Note that our approach is not limited to person re-identification with fixed camera settings. Instead, it can also be applied to capture the camera-and-person configuration and cross-view correspondence for unfixed cameras, as demonstrated in the experimental results.

In summary, our contributions to person Re-ID are three folds. First, we introduce a correspondence structure to encode cross-view correspondence pattern between cameras, and develop a global-based matching process by combining a global constraint with the correspondence structure to exclude spatial misalignments between images. These two components in fact establish a novel framework for addressing the person Re-ID problem. Second, under this framework, we propose a boosting-based approach to learn a suitable correspondence structure between a camera pair. The learned correspondence structure can not only capture the spatial correspondence pattern between cameras but also handle the viewpoint or human-pose variation in individual images. Third, this paper releases a new and challenging benchmark \textsc{Road dataset} for person Re-ID.

The rest of this paper is organized as follows. Sec.~\ref{section:related_work} reviews related works. Sec.~\ref{section:overview} describes the framework of the proposed approach. Sections~\ref{section:mapping structure learning} to~\ref{section:learning} describe the details of our proposed global-based matching process and boosting-based learning approach, respectively. Sec.~\ref{section:experimental evaluation} shows the experimental results and Sec.~\ref{section:conclusion} concludes the paper.

\section{Related Works\label{section:related_work}}

Many person re-identification methods have been proposed. Most of them focus on developing suitable feature representations about humans' appearance \cite{SalientColor, rankboost, color_in,3,OTF}, or finding proper metrics to measure the cross-view appearance similarity between images \cite{eiml,1,kernel-based,prid}. Since these works do not effectively model the spatial misalignment among local patches inside images, their performances are often limited due to the interferences from viewpoint changes and human-pose variations.

In order to address the spatial misalignment problem, some patch-based methods are proposed \cite{2,part,part2,12,6,8,10,SLEPK} which decompose images into patches and perform an online patch-level matching to exclude patch-wise misalignments. In \cite{2,part2}, a human body in an image is first parsed into semantic parts (e.g., head and torso). And then, similarity matching is performed between the corresponding semantic parts. Since these methods are highly dependent on the accuracy of body parser, they have limitations in scenarios where the body parser does not work reliably.

In \cite{part}, Oreifej et al. divide images into patches according to appearance consistencies and utilize the Earth Movers Distance (EMD) to measure the overall similarity among the extracted patches. However, since the spatial correlation among patches are ignored during similarity calculation, the method is easily affected by the mismatched patches with similar appearance. Although Ma et al. \cite{12} introduce a body prior constraint to avoid mismatching between distant patches, the problem is still not well addressed, especially for the mismatching between closely located patches.

To reduce the effect of patch-wise mismatching, some saliency-based approaches \cite{6,8} are recently proposed, which estimate the saliency distribution relationship between images and utilize it to control the patch-wise matching process. Although these methods consider the correspondence constraint between patches, our approach differs from them in: (1) our approach focuses on constructing a correspondence structure where patch-wise matching parameters are jointly decided by both matched patches. Comparatively, the matching weights in the saliency-based approach \cite{8} is only controlled by patches in the probe-image (probe patch). (2) Our approach models patch-wise correspondence by a one-to-many graph such that each probe patch will trigger multiple matches during the patch matching process. In contrast, the saliency-based approaches only select one best-matched patch for each probe patch. (3) Our approach introduces a global constraint to control the patch-wise matching result while the patch matching result in saliency-based approaches is locally decided by choosing the best-matched one within a neighborhood set.

\section{Overview}
\label{section:overview}

\begin{figure}
  \centering
  \includegraphics[width=0.475\textwidth]{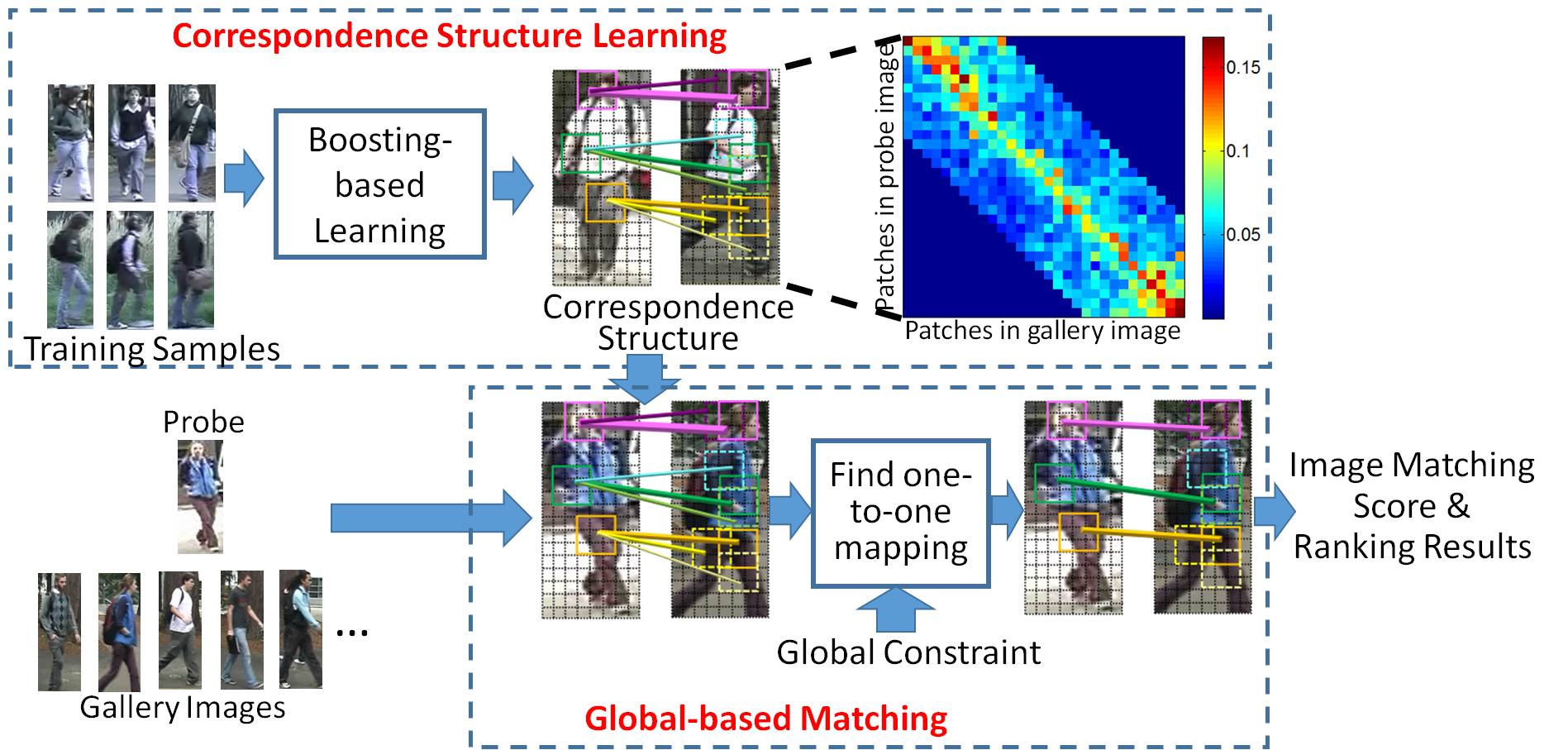}
  \caption{Framework of the proposed approach.}
    \label{fig:framework}
\end{figure}

The framework of our approach is shown in Fig.~\ref{fig:framework}.
During the training process,
which is detailed in Section~\ref{section:learning},
we present a boosting-based process
to learn the correspondence structure between the target camera pair.
During the prediction stage,
which is detailed in Section~\ref{section:mapping structure learning}
given a probe image
and a set of gallery images,
we use the correspondence structure to evaluate the patch correlations between the probe image and each gallery image,
and find the optimal one-to-one mapping between patches,
and accordingly the matching score.
The Re-ID result is achieved by ranking gallery images according to their matching scores.

\section{Person Re-Identification with Correspondence Structure\label{section:mapping structure learning}}
In this section,
we introduce the concept of correspondence structure,
show the scheme of computing the patch correlation
using the correspondence structure,
and finally present the patch-wise mapping method to
compute the matching score between the probe image and the gallery image.

\subsection{Correspondence structure}
The correspondence structure, $\mathbf{\Theta}_{A,B}$, encodes the spatial correspondence distribution
between a pair of cameras, $A$ and $B$.
In our problem, we adopt a discrete distribution,
which is a set of patch-wise matching probabilities,
$\mathbf{\Theta}_{A, B} = \{P(x^A_i, B)\}_{i=1}^{N_A}$,
where $N_A$ is the number of patches of an image in camera $A$.
$P(x^A_i, B) = \{P(x^A_i, x^B_1), P(x^A_i, x^B_2), \dots, P(x^A_i, x^B_{N_B})\}$
describes the correspondence distribution
in an image from camera $B$
for the $i$th patch $x^A_i$
of an image captured from camera $A$, where $N_B$ is the number of patches of an image in $B$.
An illustration of the correspondence distribution
is shown on the top-right
of Fig.~\ref{fig:mapping structure_example_c}.

The definition of the matching probabilities in the correspondence structure only depends on a camera pair
and are independent to the specific images.
In the correspondence structure,
it is possible that
one patch in camera $A$ is highly correlated to multiple patches in camera $B$,
so as to handle human-pose variations and local viewpoint changes in a camera view.

\subsection{Patch correlation}
Given a probe image $U$ in camera $A$ and a gallery image $V$ in camera $B$,
the patch-wise correlation between $U$ and $V$,
$C(x^U_i,x^V_j)$,
is computed from both the correspondence structure between cameras $A$ and $B$
and the visual features
and written as:
\begin{equation}
 C(x^U_i, x^V_j) = \lambda_{T_c}(P(x^U_i,x^V_j)) \cdot \log \Phi(\mathbf{f}_{x^U_i},\mathbf{f}_{x^V_j}; x^U_i, x^V_j).
\label{equation:equ1}
\end{equation}
Here $x^U_i$ and $x^V_j$ are $i$th and $j$th patch in images $U$ and $V$;
$\mathbf{f}_{x^U_i}$ and $\mathbf{f}_{x^V_j}$ are the feature vectors for $x^U_i$ and $x^V_j$.
$P(x^U_i,x^V_j) = P(x^A_i,x^B_j)$ is the correspondence structure of cameras $A$ and $B$.
$\lambda_{T_c}(P(x^U_i,x^V_j) = 1$ $P(x^U_i,x^V_j)>T_c$, and $0$ otherwise,
and $T_c=0.05$ is a threshold.
$\Phi(\mathbf{f}_{x^U_i},\mathbf{f}_{x^V_j};x^U_i, x^V_j)$ is the correspondence-structure-controlled similarity
between $x^U_i$ and $x^V_j$,
\begin{equation}
\Phi(\mathbf{f}_{x^U_i},\mathbf{f}_{x^V_j}; x^U_i, x^V_j) = \Phi_z ( \mathbf{f}_{x^U_i},\mathbf{f}_{x^V_j})  P(x^U_i,x^V_j) ,
\label{equation:equ2}
\end{equation}
where $\Phi_{z}(\mathbf{f}_{x^U_i},\mathbf{f}_{x^V_j})$ is the appearance similarity between $x^U_i$ and $x^V_j$.

The correspondence structure $P(x^U_i,x^V_j)$
in Equations~\ref{equation:equ1} and~\ref{equation:equ2},
is used to adjust the appearance similarity $\Phi_{z}(\mathbf{f}_{x^U_i},\mathbf{f}_{x^V_j})$
such that a more reliable patch-wise correlation strength can be achieved.
The thresholding term $\lambda_{T_c}(P(x^U_i,x^V_j))$ is introduced
to exclude the patch-wise correlation with a low correspondence probability,
which effectively reduces the interferences from mismatched patches with similar appearance.

The patch-wise appearance similarity $\Phi_z ( \mathbf{f}_{x^U_i},\mathbf{f}_{x^V_j})$ in Eq.~\ref{equation:equ2}
can be achieved by many off-the-shelf methods~\cite{6,8,patch_match}.
In this paper, we extract Dense SIFT and Dense Color Histogram~\cite{6} from each patch
and utilize the KISSME distance metric~\cite{1} to compute $\Phi_{z}( \Phi_z ( \mathbf{f}_{x^U_i},\mathbf{f}_{x^V_j}))$
(note that we train different KISSME metrics for patch-pairs at different locations).

\subsection{Patch-wise mapping}

\begin{figure}[t]
  \centering
  \subfloat[]{\includegraphics[width=2.9cm,height=2.95cm]{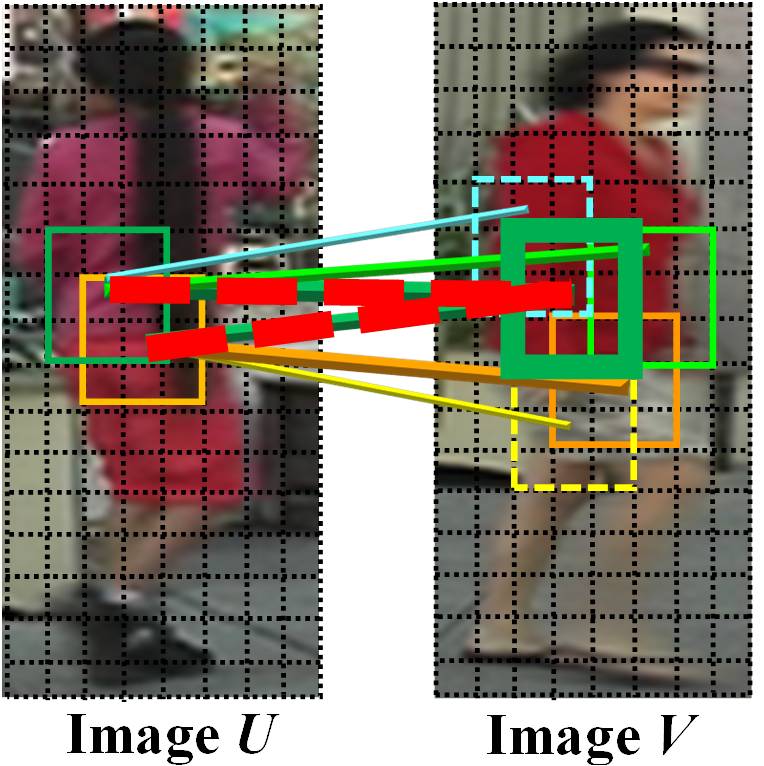}
    \label{fig:one to one matching_a}}
    \hspace{6mm}
  \subfloat[]{\includegraphics[width=2.9cm,height=2.95cm]{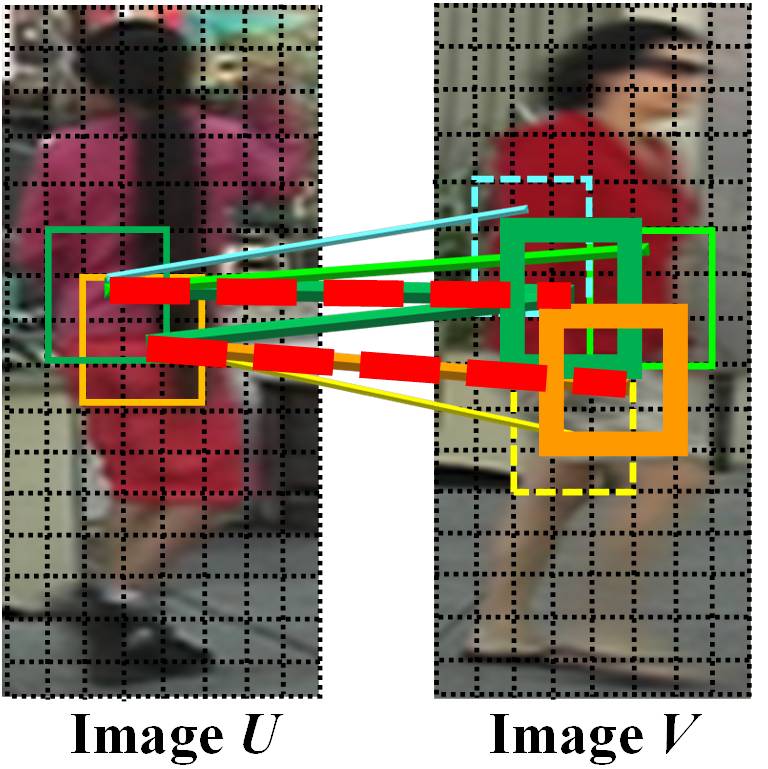}
    \label{fig:one to one matching_b}}
  \caption{Patch matching result (a) by locally finding the largest correlation strength $C(x^U_i,x^V_j)$ for each patch and (b) by using a global constraint. The red dashed lines indicate the final patch matching results and the colored solid lines are the matching probabilities in the correspondence structure. (Best viewed in color)}
  \label{fig:one to one matching}
\end{figure}

With $C(x^U_i,x^V_j)$, the alignment-enhanced correlation strength, we can find a best-matched patch in image $V$ for each patch in $U$ and herein calculate the final image matching score. However, locally finding the largest $C(x^U_i,x^V_j)$ may still create mismatches among patch pairs with high matching probabilities. For example, Fig.~\ref{fig:one to one matching_a} shows an image pair $U$ and $V$ containing different people. When locally searching for the largest $C(x^U_i,x^V_j)$, the yellow patch in $U$ will be mismatched to the bold-green patch in $V$ since they have both large appearance similarity and high matching probability. This mismatch unsuitably increases the matching score between $U$ and $V$.

To address this problem, we introduce a global one-to-one mapping constraint and solve the resulting linear assignment task~\cite{hungary} to find the best matching:
\begin{align}
&\mathbf{\Omega}^*_{U,V} = \mathop{\arg\max}_{\mathbf{\Omega}_{U,V}}{\sum_{\{x^U_i,x^V_j\}\in\mathbf{\Omega}_{U,V}}{C(x^U_i,x^V_j)}}
\label{equation:equ3}\\
\text{s.t.}~& x^U_i\neq x^U_s,x^V_j\neq x^V_t ~~~\forall~ \{x^U_i,x^V_j\}, \{x^U_s,x^V_t\}\in\mathbf{\Omega}_{U,V} \notag
\end{align}
where $\mathbf{\Omega}^*_{U,V}$ is the set of the best patch matching result between images $U$ and $V$. $\{x^U_i,x^V_j\}$ and $\{x^U_s,x^V_t\}$ are two matched patch pairs in $\mathbf{\Omega}$. According to Eq.~\ref{equation:equ3}, we want to find the best patch matching result $\mathbf{\Omega}^*_{U,V}$ that maximizes the total image matching score
\begin{align}
\psi_{U,V}=\sum_{\{x^U_i,x^V_j\}\in\mathbf{\Omega}_{U,V}}{C(x^U_i,x^V_j)}, \label{equation:matchingscore}
\end{align}
given that each patch in $U$ can only be matched to one patch in $V$ and vice versa.

Eq.~\ref{equation:equ3} can be solved by the Hungary method~\cite{hungary}. Fig.~\ref{fig:one to one matching_b} shows an example of the patch matching result by Eq.~\ref{equation:equ3}. From Fig.~\ref{fig:one to one matching_b}, it is clear that by the inclusion of a global constraint, local mismatches can be effectively reduced and a more reliable image matching score can be achieved. Based on the above process, we can calculate the image matching scores $\psi$ between a probe image and all gallery images in a cross-view camera, and rank the gallery images accordingly to achieve the final Re-ID result~\cite{12}.

\section{Correspondence Structure Learning}\label{section:learning}

\subsection{Objective function}
Given a set of probe images $\{U_\alpha\}$ from camera $A$ and
their corresponding cross-view images $\{V_\beta\}$
from camera $B$ in the training set,
we learn the optimal correspondence structure $\mathbf{\Theta}_{A,B}^*$
between cameras $A$ and $B$
so that the correct match image
is ranked before the incorrect match images
in terms of the matching scores.
The formulation is give as below,
\begin{align}
  \mathop{\min}_{\mathbf{\Theta}_{A,B}}
    \sum_{U_{\alpha}} R(V_{\alpha'}; \psi_{U_\alpha,V_{\alpha'}}(\mathbf{\Theta}_{A,B}),  \mathbf{\Psi}_{U_\alpha,V_{\beta\neq\alpha'}} (\mathbf{\Theta}_{A,B})), \label{equation:equ4}
\end{align}
where $V_{\alpha'}$ is the correct match gallery image of the probe image ${U_{\alpha}}$.
$\psi_{U_\alpha,V_{\alpha'}}(\mathbf{\Theta}_{A,B})$ (as computed from Eq.~\ref{equation:matchingscore})
is the matching score
between $U_\alpha$ and $V_{\alpha'}$
and $\mathbf{\Psi}_{U_\alpha,V_{\beta\neq\alpha'}} (\mathbf{\Theta}_{A,B})$
is the set of matching scores of all incorrect match images.
$R(V_{\alpha'}; \psi_{U_\alpha,V_{\alpha'}}(\mathbf{\Theta}_{A,B}),  \mathbf{\Psi}_{U_\alpha,V_{\beta\neq\alpha'}} (\mathbf{\Theta}_{A,B}))$
is the rank of $V_{\alpha'}$
among all the gallery images
according to the matching scores.
Intuitively,
the penalty is the smallest
if the rank is $1$,
i.e., the matching score of $V_{\alpha'}$
is the greatest.
The optimization is not easy
as the matching score calculation (Eq.~\ref{equation:matchingscore}) is complicated.
We present an approximate solution, a boosting-based process,
to solve this problem.

\subsection{Boosting-based learning}\label{section:boost-learning}

\begin{figure}[t]
  \centering
  \subfloat[]{\includegraphics[width=2.37cm,height=2.4cm]{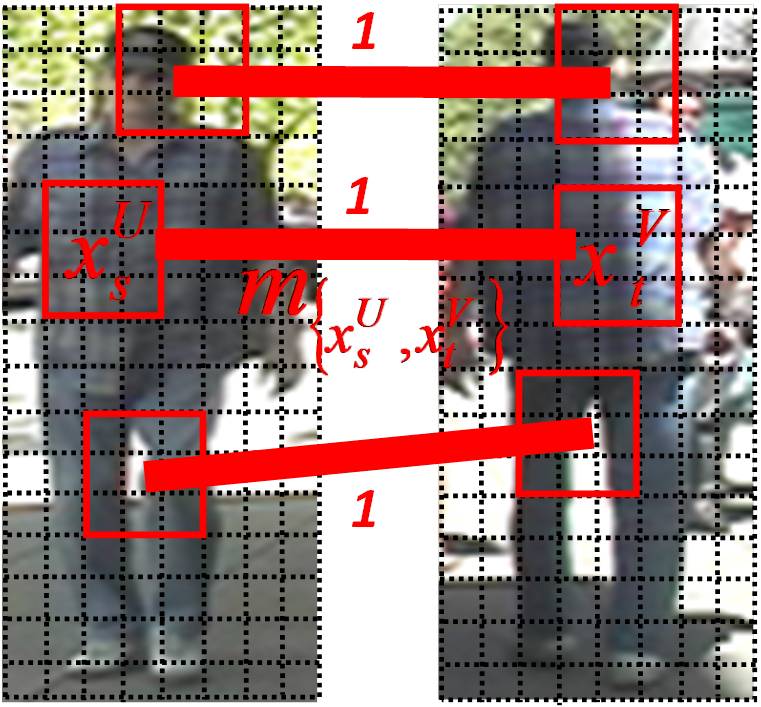}
    \label{fig:one to many matching a}}
  \hspace{1mm}
  \subfloat[]{\includegraphics[width=2.4cm,height=2.4cm]{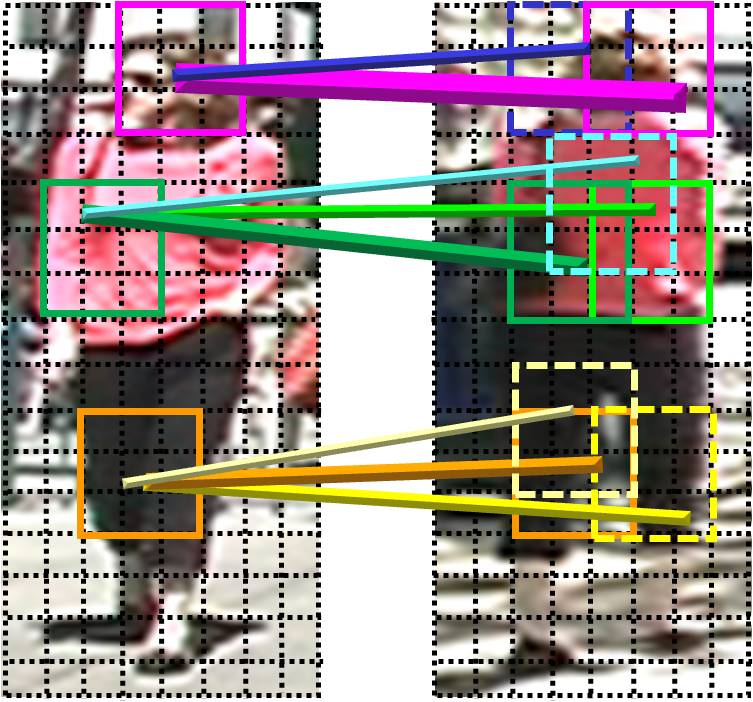}
    \label{fig:one to many matching b}}
  \hspace{1mm}
  \subfloat[]{\includegraphics[width=2.4cm,height=2.4cm]{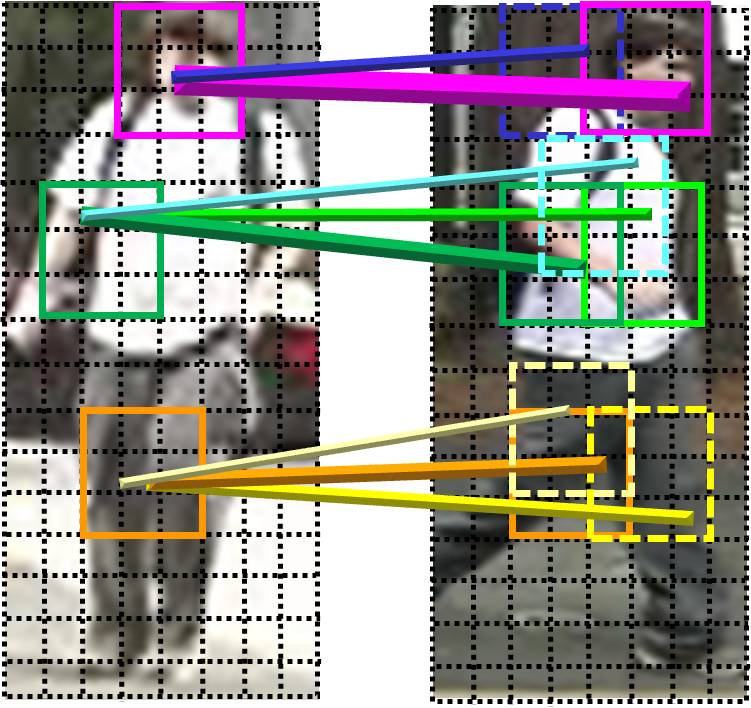}
    \label{fig:one to many matching c}}
  \\
  \subfloat[]{\includegraphics[width=2.4cm,height=2.4cm]{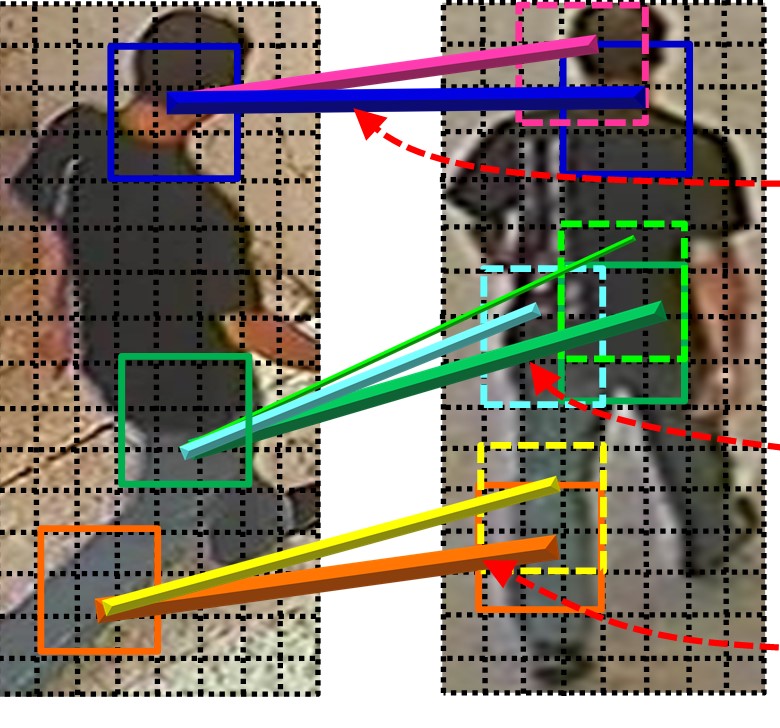}
    \label{fig:one to many matching d}}
  \hspace{-1.8mm}
  \subfloat[]{\includegraphics[width=2.59cm,height=2.4cm]{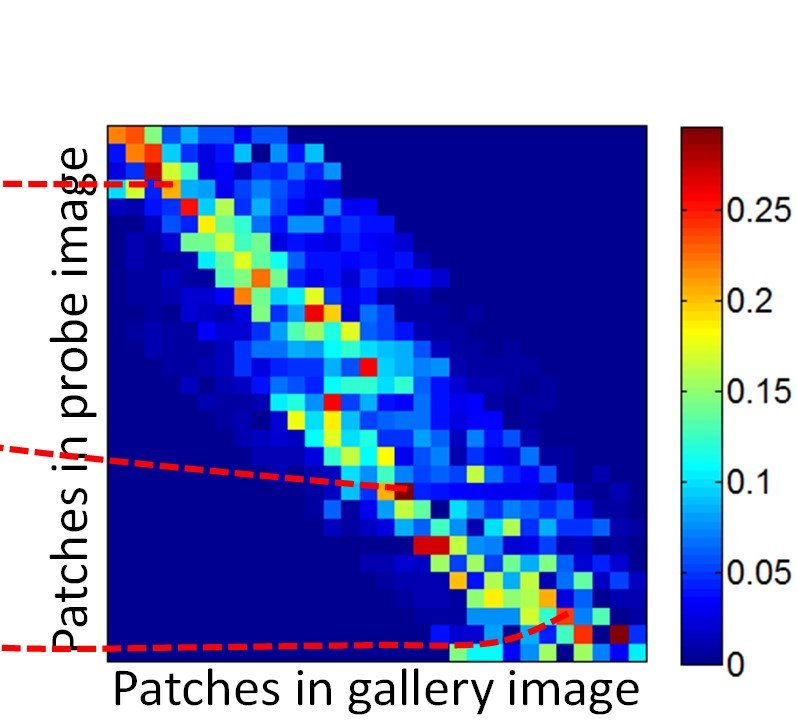}
    \label{fig:one to many matching e}}
  \hspace{1mm}
  \subfloat[]{\includegraphics[width=2.46cm,height=2.04cm]{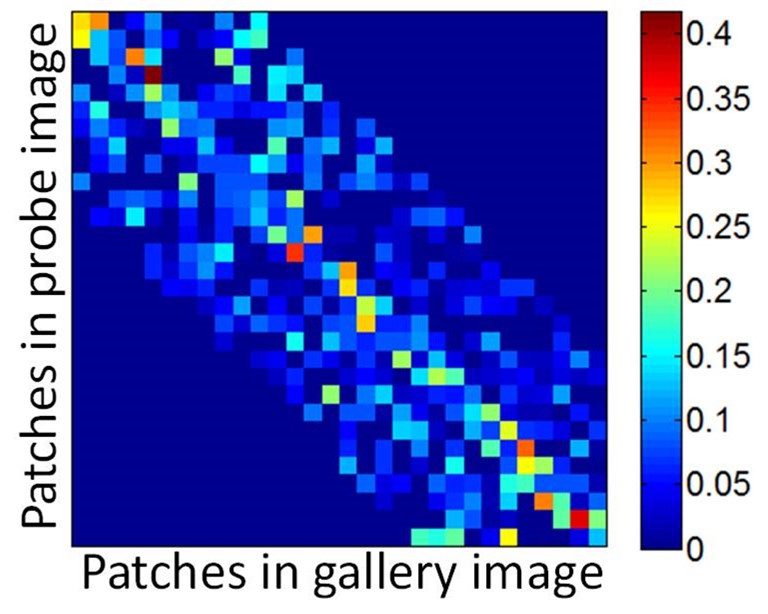}
    \label{fig:one to many matching f}}
  \caption{(a): An example of binary mapping structure (the red lines with weight $1$ indicate that the corresponding patches are connected). (b)-(d): Examples of the correspondence structures learned by our approach where (b)-(c) and (d) are the correspondence structures for the VIPeR~\cite{viper} and 3DPeS~\cite{3dpes} datasets, respectively. The line widths in (b)-(d) are proportional to the patch-wise probability values. (e): The complete correspondence structure matrix of (d) learned by our approach. (f): The correspondence structure matrix of (d)'s dataset obtained by the simple-average method. (Patches in (e) and (f) are organized by a zig-zag scanning order. Matrices in (e) and (f) are down-sampled for a clearer illustration of the correspondence pattern). (Best viewed)
  }
    \label{fig:one to many matching}
\end{figure}

The boosting-based approach utilizes a progressive way to find the best correspondence structure with the help of \emph{binary mapping structures}. A binary mapping structure is similar to the correspondence structure except that it simply utilizes $0$ or $1$ instead of matching probabilities to indicate the connectivity or linkage between patches, cf. Fig.~\ref{fig:one to many matching a}. It can be viewed as a simplified version of the correspondence structure which includes rough information about the cross-view correspondence pattern.

Since binary mapping structures only include simple connectivity information among patches, their optimal solutions are tractable for individual probe images. Therefore, by searching for the optimal binary mapping structures for different probe images and utilizing them to progressively update the correspondence structure, suitable cross-view correspondence patterns can be achieved.

The entire boosting-based learning process can be describe by the following steps as well as Algorithm~\ref{algorithm:progressive updating}.

{\bf Finding the optimal binary mapping structure.} For each training probe image $U_{\alpha}$, we first create multiple candidate binary mapping structures under different search ranges by adjacency-constrained search ~\cite{6}, and then find the optimal binary mapping structure $\mathbf{M}_\alpha$ such that the rank order of $U_{\alpha}$'s correct match image $V_{\alpha'}$ is minimized under $\mathbf{M}_\alpha$. Note that we find one optimal binary mapping structure for each probe image such that the obtained binary mapping structures can include local cross-view correspondence information in different training samples.

{\bf Correspondence Structure Initialization.} In this paper, patch-wise matching probabilities $P(x^U_i,x^V_j)$ in the correspondence structure are initialized by:
\begin{align}
        \label{equation:equ5a}
        P^0(x^U_i,x^V_j)
        \propto\left\{
           \begin{aligned}
           &0,~~~~~~~~~~~~\text{if}~~~d{(x^V_i,x^V_j)}\geq T_d\\
           &\frac {1}{d{(x^V_i,x^V_j)}+1},~~~~~\text{otherwise}
           \end{aligned}
        \right.\,,
\end{align}
where $x^V_i$ is {$x^U_i$}'s co-located patch in camera $B$. $d{(x^V_i,x^V_j)}$ is the distance between patches $x^V_i$ and $x^V_j$. It is defined as the number of strides to move from $x^V_i$ to $x^V_j$ in the zig-zag order. $T_d$ is a threshold which is set to be $32$ in this paper. According to Eq.~\ref{equation:equ5a}, $P^0(x^U_i,x^V_j)$ is inversely proportional to the co-located distance between $x^V_i$ and $x^V_j$ and will equal to $0$ if the distance is larger than a threshold.

{\bf Binary mapping structure selection.} During each iteration $k$ in the learning process, we first apply correspondence structure $\mathbf{\Theta}^{k-1}_{A,B}=\{P^{k-1}(x^U_i,x^V_j)\}$ from the previous iteration to calculate the rank orders of all correct match images $V_{\alpha'}$ in the training set. Then, we randomly select $20$ $V_{\alpha'}$ where half of them are ranked among top $50\%$ (implying better Re-ID results) and another half are ranked among the last $50\%$ (implying worse Re-ID results). Finally, we extract binary mapping structures corresponding to these selected images and utilize them to update and boost the correspondence structure.

Note that we select binary mapping structures for both high- and low-ranked images in order to include a variety of local patch-wise correspondence patterns. In this way, the final obtained correspondence structure can suitably handle the variations in human-pose or local viewpoints.

\begin{algorithm}[t]
   \caption{Boosting-based Learning Process}
   \small{
     {\bf Input}: A set of training probe images $\{U_\alpha\}$ from camera $A$ and their corresponding cross-view images $\{V_\beta\}$ from camera $B$\\
     {\bf Output}: $\mathbf{\Theta}_{A,B}=\{P(x^U_i,y^V_j)\}$, the correspondence structure between $\{U_\alpha\}$ and $\{V_\beta\}$
   }
     \begin{algorithmic}[1]
     \small{
      \STATE Find an optimal binary mapping structure $\mathbf{M}_\alpha$ for each probe image $U_{\alpha}$, as described in the $4$-th paragraph in Sec~\ref{section:boost-learning}
      \STATE Set $k=1$. Initialize $P^0(x^U_i,y^V_j)$ by Eq.~\ref{equation:equ5a}.
      \STATE Use the current correspondence structure $\{P^{k-1}{(x^U_i,x^V_j)}\}$ to perform Re-ID on $\{U_\alpha\}$ and $\{V_\beta\}$, and select $20$ binary mapping structures $\mathbf{M}_\alpha$ based on the Re-ID result, as described in the $6$-th paragraph in Sec~\ref{section:boost-learning}
      \STATE Compute updated match probability $\hat{P}^k{(x^U_i,x^V_j)}$ by Eq.~\ref{equation:equ5}
      \STATE Update the matching probabilities $P^k{(x^U_i,x^V_j)}$ by Eq.~\ref{equation:equ11}
      \STATE Set $k=k+1$ and go back to step $3$ if not converged or not reaching the maximum iteration number
      \STATE Output $\{P(x^U_i,y^V_j)\}$
     }
     \end{algorithmic}
     \label{algorithm:progressive updating}
\end{algorithm}

{\bf Calculating the updated matching probability.} With the introduction of the binary mapping structure $\mathbf{M}_\alpha$, we can model the updated matching probability in the correspondence structure by:
\begin{equation}
\hat{P}^k{(x^U_i,x^V_j)}=\sum_{\mathbf{M}_\alpha\in\mathbf{\Gamma}^k}{\hat{P}(x^U_i,x^V_j|\mathbf{M_\alpha})\cdot P(\mathbf{M}_\alpha)} \,,
\label{equation:equ5}
\end{equation}
where $\hat{P}^k{(x^U_i,x^V_j)}$ is the updated matching probability between patches $x^U_i$ and $x^V_j$ in the $k$-th iteration. $\mathbf{\Gamma}^k$ is the set of binary mapping structures selected in the $k$-th iteration. $P(\mathbf{M}_\alpha)=\frac{\tilde{\mathcal{R}}_n(\mathbf{M}_\alpha)}{\sum_{\mathbf{M}_\gamma\in\mathbf{\Gamma}^k}\tilde{\mathcal{R}}_n(\mathbf{M}_\gamma)}$ is the prior probability for binary mapping structure $\mathbf{M}_\alpha$, where $\tilde{\mathcal{R}}_n(\mathbf{M}_\alpha)$ is the CMC score at rank $n$~\cite{cmc} when using $\mathbf{M}_\alpha$ as the correspondence structure to perform person Re-ID over the training images. $n$ is set to be $5$ in our experiments.

$\hat{P}(x^U_i,x^V_j|\mathbf{M_\alpha})$ is the updated matching probability between $x^U_i$ and $x^V_j$ when including the local correspondence pattern information of $\mathbf{M}_\alpha$. It can be calculated by:
\begin{equation}
\hat{P}(x^U_i,x^V_j|\mathbf{M}_\alpha)=\hat{P}(x^V_j|x^U_i,\mathbf{M}_\alpha)\cdot \hat{P}(x^U_i|\mathbf{M}_\alpha) \,, \label{equation:equ6}
\end{equation}
$\hat{P}(x^V_j|x^U_i,\mathbf{M}_\alpha)$ is the updated probability to correspond from $x^U_i$ to $x^V_j$ when including $\mathbf{M}_\alpha$, calculated as
\begin{align}
\label{equation:equ7}
        \hat{P}(x^V_j|x^U_i,\mathbf{M}_\alpha)\propto
        \left\{
           \begin{aligned}
           &1,~~~~~\text{if}~~m_{\{x^U_i,x^V_j\}}\in\mathbf{M}_\alpha \\
           &\tilde{\mathcal{A}}_{x^V_j|x^U_i,\mathbf{M}_\alpha},~~~\text{otherwise}
           \end{aligned}
        \right. \,,
\end{align}
where $m_{\{x^U_i,x^V_j\}}$ is a patch-wise link connecting $x^U_i$ and $x^V_j$. $\tilde{\mathcal{A}}_{x^V_j|x^U_i,\mathbf{M}_\alpha}=\frac{\overline{\Phi}_z(x^U_i,x^V_j)}{\sum_{x^V_t,m_{\{x^U_i,x^V_t\}}\in\mathbf{M}_\alpha}{\overline{\Phi}_z(x^U_i,x^V_t)}}$, where $\overline{\Phi}_z(x^U_i,x^V_j)$ is the average appearance similarity~\cite{6,1} between patches $x^U_i$ and $x^V_j$ over all correct match image pairs in the training set. $x^V_t$ is a patch that is connected to $x^U_i$ in the binary mapping structure $\mathbf{M}_\alpha$. From Eq.~\ref{equation:equ7}, $\hat{P}(x^V_j|x^U_i,\mathbf{M}_\alpha)$  will equal to $1$ if $\mathbf{M}_\alpha$ includes a link between $x^U_i$ and $x^V_j$. Otherwise, $\hat{P}(x^V_j|x^U_i,\mathbf{M}_\alpha)$ will be decided by the relative appearance similarity strength between patch pair $\{x^U_i, x^V_j\}$ and all patch pairs which are connected to $x^U_i$ in the binary mapping structure $\mathbf{M}_\alpha$.

Furthermore, $\hat{P}(x^U_i|\mathbf{M}_\alpha)$ in Eq.~\ref{equation:equ6} is the updated importance probability of $x^U_i$ after including $\mathbf{M}_\alpha$. It can be calculated by integrating the importance probability of each individual link in $\mathbf{M}_\alpha$:
\begin{align}
 \label{equation:equ8}
 \hat{P}(x^U_i|\mathbf{M}_\alpha)=\sum_{m_{\{x^U_s,x^V_t\}}\in\mathbf{M}_\alpha}{}&{\hat{P}(x^U_i|m_{\{x^U_s,x^V_t\}},\mathbf{M}_\alpha)}\notag\\
 &\cdot\hat{P}(m_{\{x^U_s,x^V_t\}}|\mathbf{M}_\alpha) \,,
\end{align}
where $m_{\{x^U_s,x^V_t\}}$ is a patch-wise link in $\mathbf{M}_\alpha$, as the red lines in Fig.~\ref{fig:one to many matching a}. $\hat{P}(m_{\{x^U_s,x^V_t\}}|\mathbf{M}_\alpha)$ is
the importance probability of link $m_{\{x^U_s,x^V_t\}}$ which is defined similar to $P(\mathbf{M}_\alpha)$:
\begin{equation}
 \hat{P}(m_{\{x^U_s,x^V_t\}}|\mathbf{M}_\alpha)=\frac{\tilde{\mathcal{R}}_n (m_{\{x^U_s,x^V_t\}})}{\sum_{m_{\{x^U_h,x^V_g\}}\in\mathbf{M}_\alpha}\tilde{\mathcal{R}}_n (m_{\{x^U_h,x^V_g\}})} \,,
\label{equation:equ9}
\end{equation}
where $\tilde{\mathcal{R}}_n (m_{\{x^U_s,x^V_t\}} )$ is the rank-$n$ CMC
score \cite{cmc} when only using a single link $m_{\{x^U_s,x^V_t\}}$ as the correspondence structure to perform Re-ID.

$\hat{P}(x^U_i|m_{\{x^U_s,x^V_t\}}, \mathbf{M}_\alpha)$ in Eq.~\ref{equation:equ8} is the impact probability from link $m_{\{x^U_s,x^V_t\}}$ to patch $x^U_i$, defined as:
\begin{align}\notag
        \label{equation:equ10}
        \hat{P}(x^U_i|m_{\{x^U_s,x^V_t\}}, \mathbf{M}_\alpha)
        \propto\left\{
           \begin{aligned}
              &0,~~~~~~~~~~\text{if } d{(x^U_i,x^U_s)}\geq T_d\\
              &\frac {1}{d{(x^U_i,x^U_s)}+1},~~\text{otherwise}
           \end{aligned}
        \right.
\end{align}
where $x^U_s$ is link $m_{\{x^U_s,x^V_t\}}$'s end patch in camera A. $d(\cdot)$ and $T_d$ are the same as Eq.~\ref{equation:equ5a}.

{\bf Correspondence structure update.} With the updated matching probability $\hat{P}^k{(x^U_i,x^V_j)}$ in Eq.~\ref{equation:equ5}, the matching probabilities in the $k$-th iteration can be finally updated by:
\begin{equation}
P^k{(x^U_i,x^V_j)} = (1-\varepsilon)P^{k-1}{(x^U_i,x^V_j)}+\varepsilon\hat{P}^k{(x^U_i,x^V_j)} \,,
\label{equation:equ11}
\end{equation}
where $P^{k-1}{(x^U_i,x^V_j)}$ is the matching probability in iteration $k-1$. $\varepsilon$ is the update rate which is set $0.2$ in our paper.

From Equations~\ref{equation:equ5}--\ref{equation:equ11}, our update process integrates multiple variables (i.e., binary mapping structure, individual links, patch-link correlation) into a unified probability framework. In this way, various information cues such as appearances, ranking results, and patch-wise correspondence patterns can be effectively included during the model updating process. Besides, although the exact convergence of our learning process is difficult to analyze due to the inclusion of rank score calculation, our experiments show that most correspondence structures become stable within $300$ iterations, which implies the reliability of our approach.

Figures~\ref{fig:mapping structure_example} and~\ref{fig:one to many matching} show some examples of the correspondence structures learned from different cross-view datasets. From Figures~\ref{fig:mapping structure_example} and~\ref{fig:one to many matching}, we can see that the correspondence structures learned by our approach can suitably indicate the matching correspondence between spatial misaligned patches. For example, in Figures~\ref{fig:mapping structure_example} and~\ref{fig:one to many matching d}-\ref{fig:one to many matching e}, the large lower-to-upper misalignments between cameras are effectively captured. Besides, the matching probability values in the correspondence structure also suitably reflects the correlation strength between different patch locations, as displayed by the colored points in Figures~\ref{fig:mapping structure_example_c} and~\ref{fig:one to many matching e}.

Furthermore, comparing Figures~\ref{fig:mapping structure_example_a} and~\ref{fig:mapping structure_example_b}, we can see that the human-pose variation is also suitably handled by the learned correspondence structure. More specifically, although images in Fig~\ref{fig:mapping structure_example} have different human poses, patches of camera $A$ in both figures can correctly find their corresponding patches in camera $B$ since the one-to-many matching probability graphs in the correspondence structure suitably embed the local correspondence variation between cameras. Similar observations can also be obtained from Figures~\ref{fig:one to many matching b} and~\ref{fig:one to many matching c}. It should be noted that images in the dataset of Figures~\ref{fig:one to many matching b} and~\ref{fig:one to many matching c} are taken by unfixed cameras (i.e., cameras with unfixed locations). However, the correspondence structure learned by our approach can still effectively encode the camera-person configuration and capture the cross-view correspondence pattern accordingly.

\section{Experimental Results\label{section:experimental evaluation}}

We perform experiments on the following four datasets:

{\bf VIPeR.} The VIPeR dataset~\cite{viper} is a commonly used dataset which contains $632$ image pairs for $632$ pedestrians, as in Figures~\ref{fig:one to many matching a}-\ref{fig:one to many matching c} and~\ref{fig:different mapping structures d}. It is one of the most challenging datasets which includes large differences in viewpoint, pose, and illumination between two camera views. Images from camera $A$ are mainly captured from $0$ to $90$ degree while camera $B$ mainly from $90$ to $180$ degree.

{\bf PRID 450S.} The PRID 450S dataset~\cite{prid} consists of $450$ person image pairs from two non-overlapping camera views. It is also challenging due to low image qualities and viewpoint changes.

{\bf 3DPeS.} The 3DPeS dataset~\cite{3dpes} is comprised of $1012$ images from $193$ pedestrians captured by eight cameras, where each person has $2$ to $26$ images, as in Figures~\ref{fig:one to many matching d} and~\ref{fig:different mapping structures a}. Note that since there are eight cameras with significantly different views in the dataset, in our experiments, we group cameras with similar views together and form three camera groups. Then, we train a correspondence structure between each pair of camera groups. Finally, three correspondence structures are achieved and utilized to perform Re-ID between different camera groups. For images from the same camera group, we simply utilize adjacency-constrained search~\cite{6} to find patch-wise mapping and calculate the image matching score accordingly.

{\bf Road.} The road dataset is our own constructed dataset which includes $416$ image pairs taken by two cameras with camera $A$ monitoring an exit region and camera $B$ monitoring a road region, as in Figures~\ref{fig:mapping structure_example} and~\ref{fig:different mapping structures g}.\footnote{This dataset will be open to the public soon.} Since images in this dataset are taken from a realistic crowd road scene, the interferences from severe occlusion and large pose variation significantly increase the difficulty of this dataset.

For all of the above datasets, we follow previous methods \cite{3,kernel-based,SalientColor} and perform experiments under $50\%$-training and $50\%$-testing. All images are scaled to $128\times 48$. The patch size in our approach is $24\times 18$. The stride size between neighboring patches is $6$ horizontally and $8$ vertically for probe images, and $3$ horizontally and $4$ vertically for gallery images. Note that we use smaller stride size in gallery images in order to obtain more patches. In this way, we can have more flexibilities during patch-wise matching.

\subsection{Results for patch matching \label{section:significance of mapping structure}}

We compare the patch matching results of three methods: (1) The adjacency-constrained search method~\cite{6,8} which finds a best matched patch for each patch in a probe image (probe patch) by searching a fixed neighborhood region around the probe patch's co-located patch in a gallery image (\emph{Adjacency-constrained}). (2) The simple-average method which simply averages the binary mapping structures for different probe images (as in Fig.~\ref{fig:one to many matching a}) to be the correspondence structure and combines it with a global constraint to find the best one-to-one patch matching result (\emph{Simple-average}). (3) Our approach which employs a boosting-based process to learn the correspondence structure and combines it with a global constraint to find the best one-to-one patch matching result.

\begin{figure}[t]
  \centering
  \subfloat[]{\includegraphics[width=2.3cm,height=2.2cm]{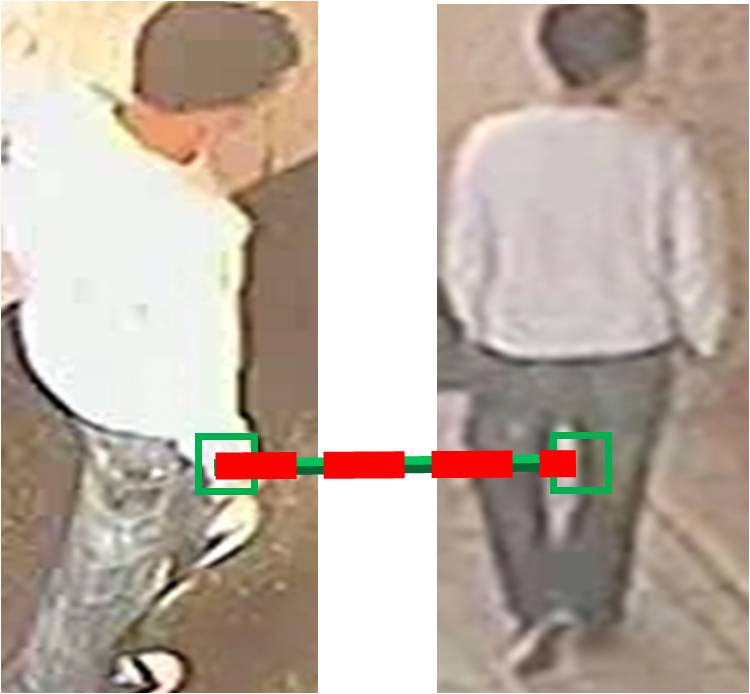}
    \label{fig:different mapping structures a}}
    \hspace{2mm}
  \subfloat[]{\includegraphics[width=2.3cm,height=2.2cm]{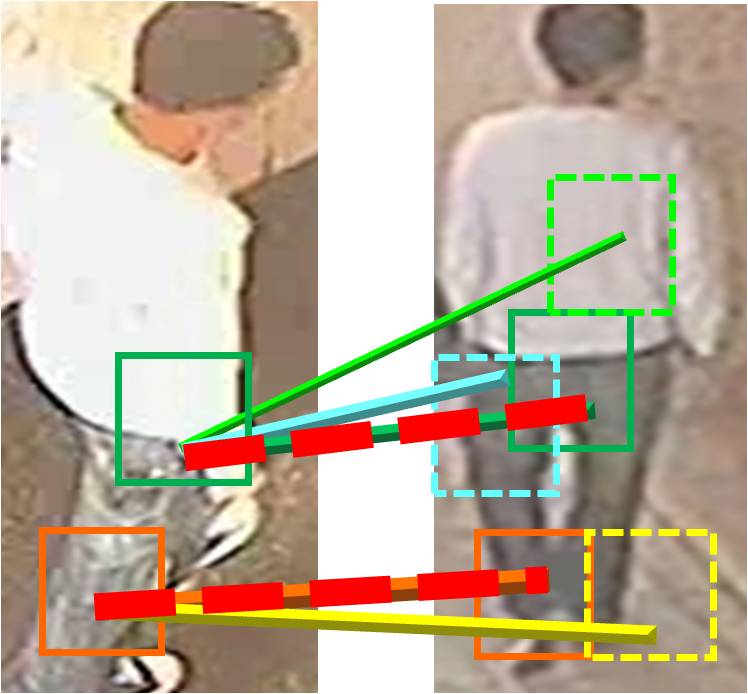}
    \label{fig:different mapping structures b}}
    \hspace{2mm}
  \subfloat[]{\includegraphics[width=2.3cm,height=2.2cm]{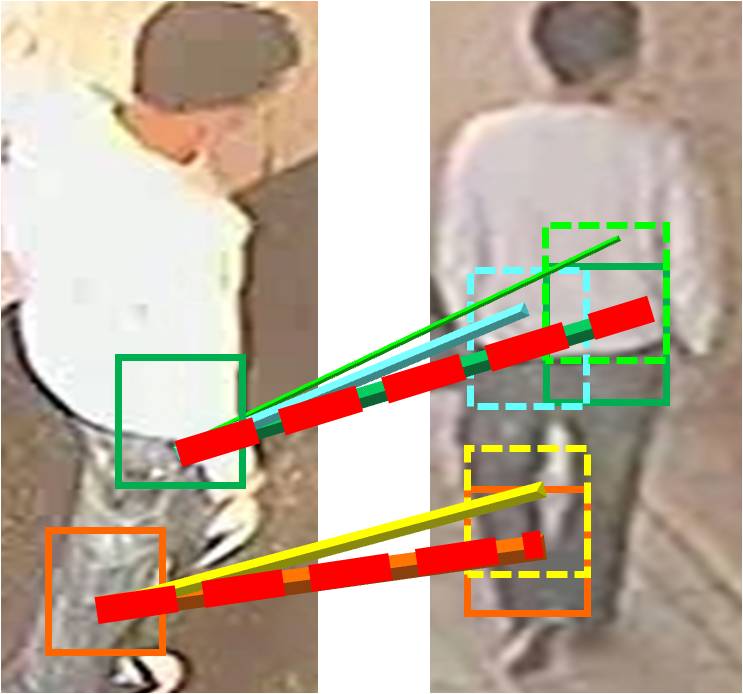}
    \label{fig:different mapping structures c}}\\
    \subfloat[]{\includegraphics[width=2.4cm,height=2.2cm]{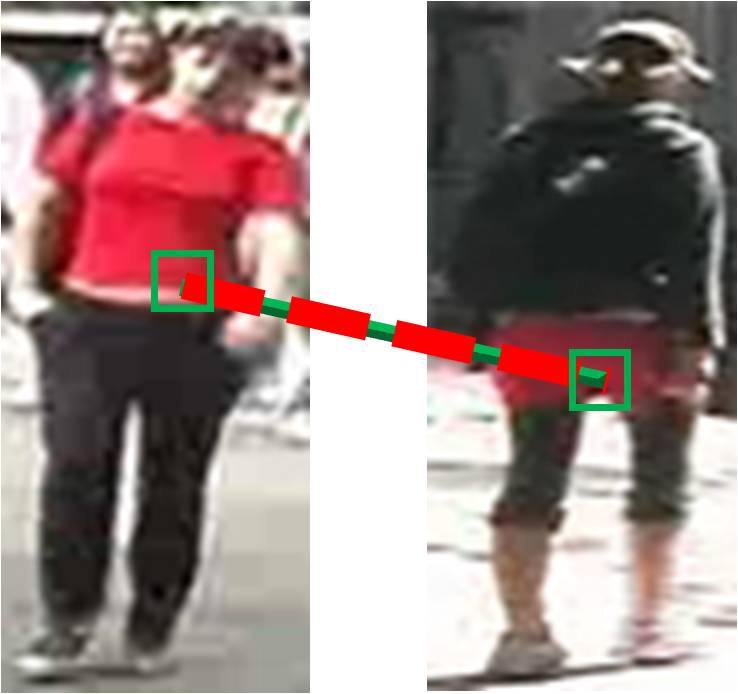}
    \label{fig:different mapping structures d}}
    \hspace{2mm}
  \subfloat[]{\includegraphics[width=2.3cm,height=2.2cm]{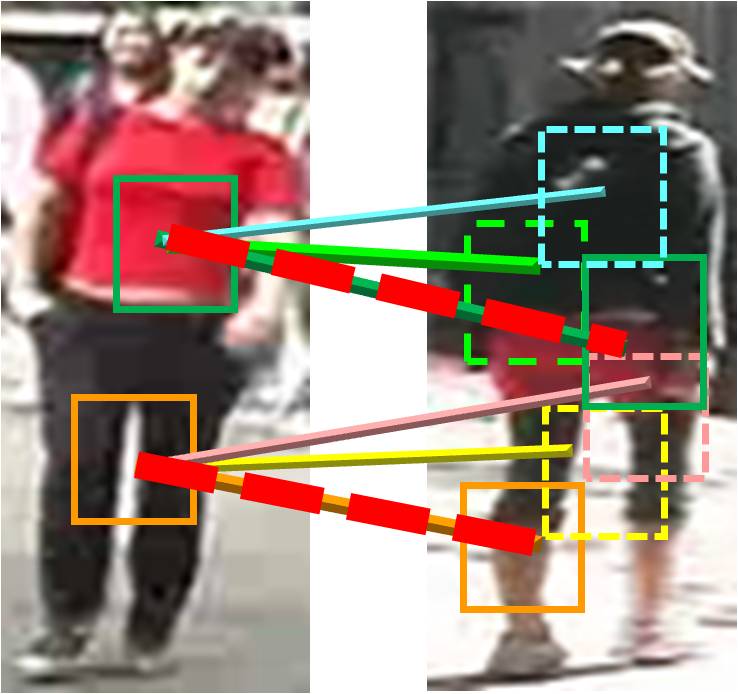}
    \label{fig:different mapping structures e}}
    \hspace{2mm}
  \subfloat[]{\includegraphics[width=2.3cm,height=2.2cm]{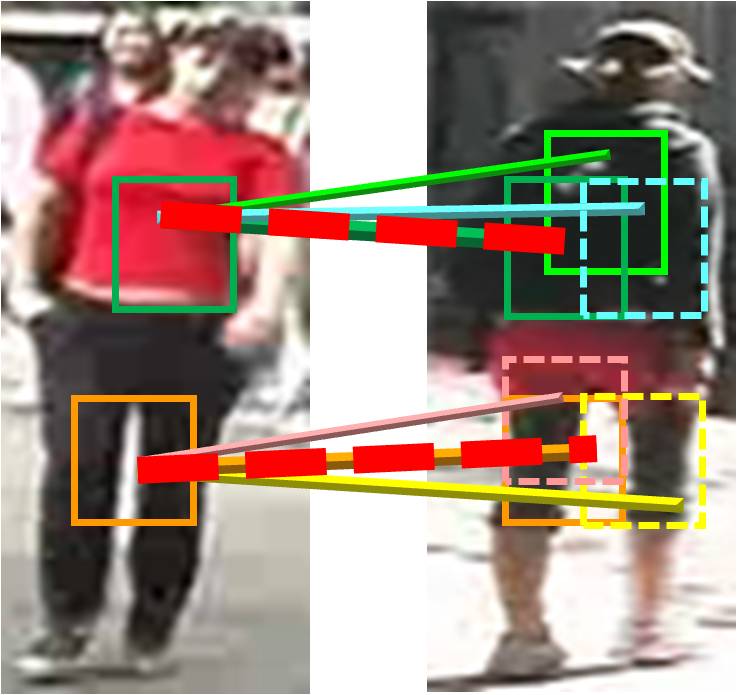}
    \label{fig:different mapping structures f}}\\
  \subfloat[]{\includegraphics[width=2.3cm,height=2.2cm]{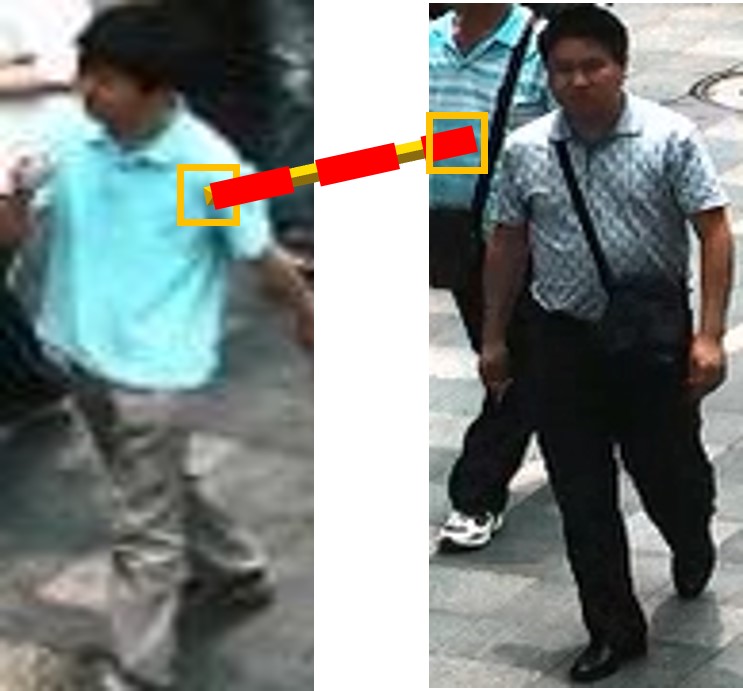}
    \label{fig:different mapping structures g}}
    \hspace{2mm}
  \subfloat[]{\includegraphics[width=2.3cm,height=2.2cm]{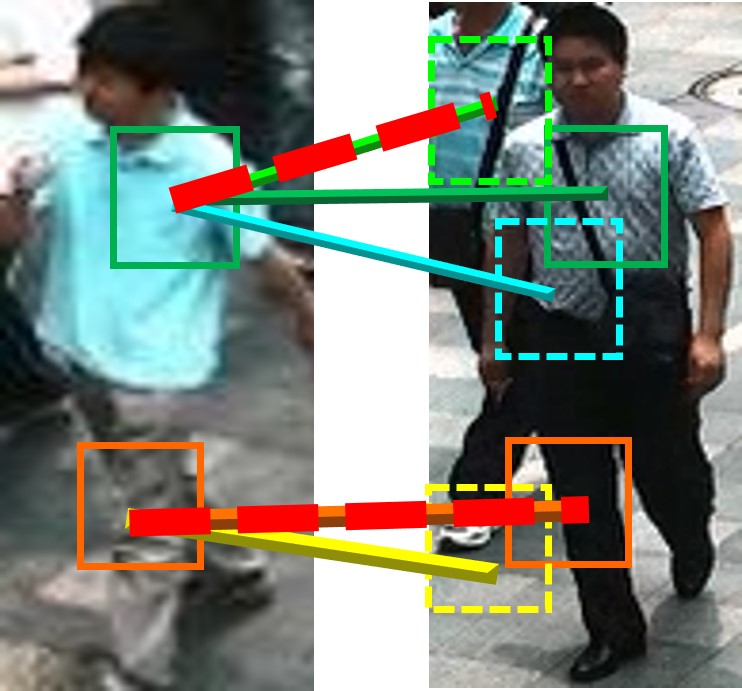}
    \label{fig:different mapping structures h}}
    \hspace{2mm}
  \subfloat[]{\includegraphics[width=2.3cm,height=2.2cm]{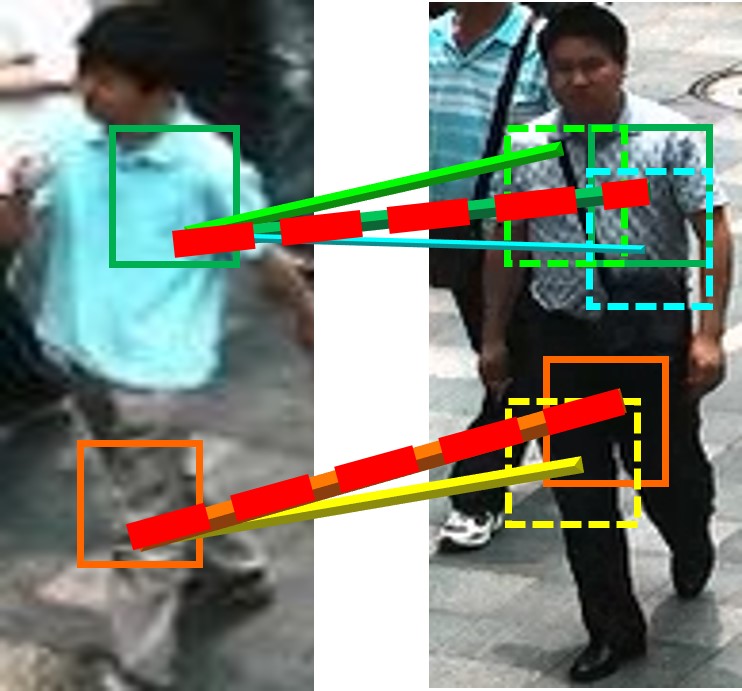}
    \label{fig:different mapping structures i}}
  \caption{Comparison of different patch mapping methods. Left column: the adjacency-constrained method; Middle column: the simple-average method; Last column: our approach. The solid lines represent matching probabilities in a correspondence structure and the red-dashed lines represent patch matching results. Note that the image pair in (a)-(c) includes the same person (i.e., correct match) while the image pairs in (d)-(i) include different people (i.e., wrong match). (Best viewed in color)}
    \label{fig:different mapping structures}
\end{figure}

Fig.~\ref{fig:different mapping structures} shows the patch mapping results of different methods, where solid lines represent matching probabilities in a correspondence structure and red-dashed lines represent patch matching results. Besides, Figures~\ref{fig:one to many matching e} and~\ref{fig:one to many matching f} show one example of the correspondence structure matrix obtained by our approach and the simple-average method, respectively. From Figures~\ref{fig:different mapping structures} and~\ref{fig:one to many matching e}-\ref{fig:one to many matching f}, we can observe:

(1) Since the adjacency-constrained method searches a fixed neighborhood region without considering the correspondence pattern between cameras, it may easily be interfered by wrong patches with similar appearances in the neighborhood (cf. Figures.~\ref{fig:different mapping structures d},~\ref{fig:different mapping structures g}). Comparatively, with the indicative matching probability information in the correspondence structure, the interference from mismatched patches can be effectively reduced (cf. Figures.~\ref{fig:different mapping structures f},~\ref{fig:different mapping structures i}).

(2) When there are large misalignments between cameras, the adjacency-constrained method may fail to find proper patches as the correct patches may be located outside the neighborhood region, as in Fig.~\ref{fig:different mapping structures a}. Comparatively, the large misalignment pattern between cameras can be properly captured by our correspondence structure, resulting in a more accurate patch matching result (cf. Fig.~\ref{fig:different mapping structures c}).

(3) Comparing Figures~\ref{fig:one to many matching e}, \ref{fig:one to many matching f} with the last two columns in Fig.~\ref{fig:different mapping structures}, it is obvious that the correspondence structures by our approach is better than the simple average method. Specifically, the correspondence structures by the simple average method include many unsuitable matching probabilities which may easily result in wrong patch matches. In contrast, the correspondence structures by our approach are more coherent with the actual spatial correspondence pattern between cameras. This implies that reliable correspondence structure cannot be easily achieved without suitably integrating the information cues between cameras.

\subsection{Results for person re-identification}

We evaluate person re-identification results by the standard Cumulated Matching Characteristic (CMC) curve~\cite{cmc} which measures the correct match rates
within different Re-ID rank ranges. The evaluation protocols are the same as~\cite{3}. That is, for each dataset, we perform $10$ randomly-partitioned $50\%$-training and $50\%$-testing experiments and average the results.

\begin{figure*}
  \centering
  \subfloat[the VIPeR dataset]{\includegraphics[width=5.3cm,height=4.3cm]{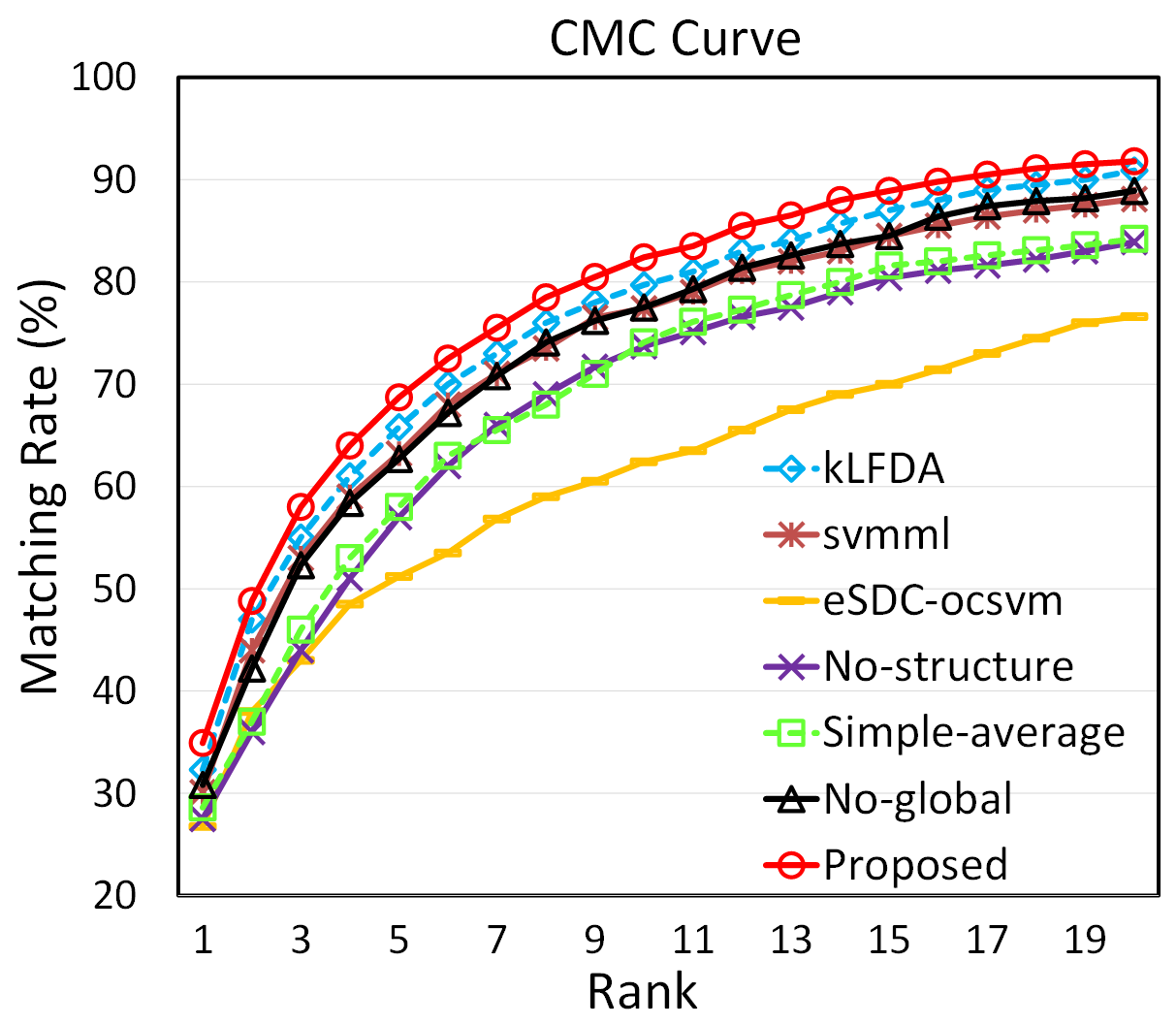}
    \label{fig:cmc curve a}}
    \hspace{3mm}
  \subfloat[the PRID 450S dataset]{\includegraphics[width=5.3cm,height=4.3cm]{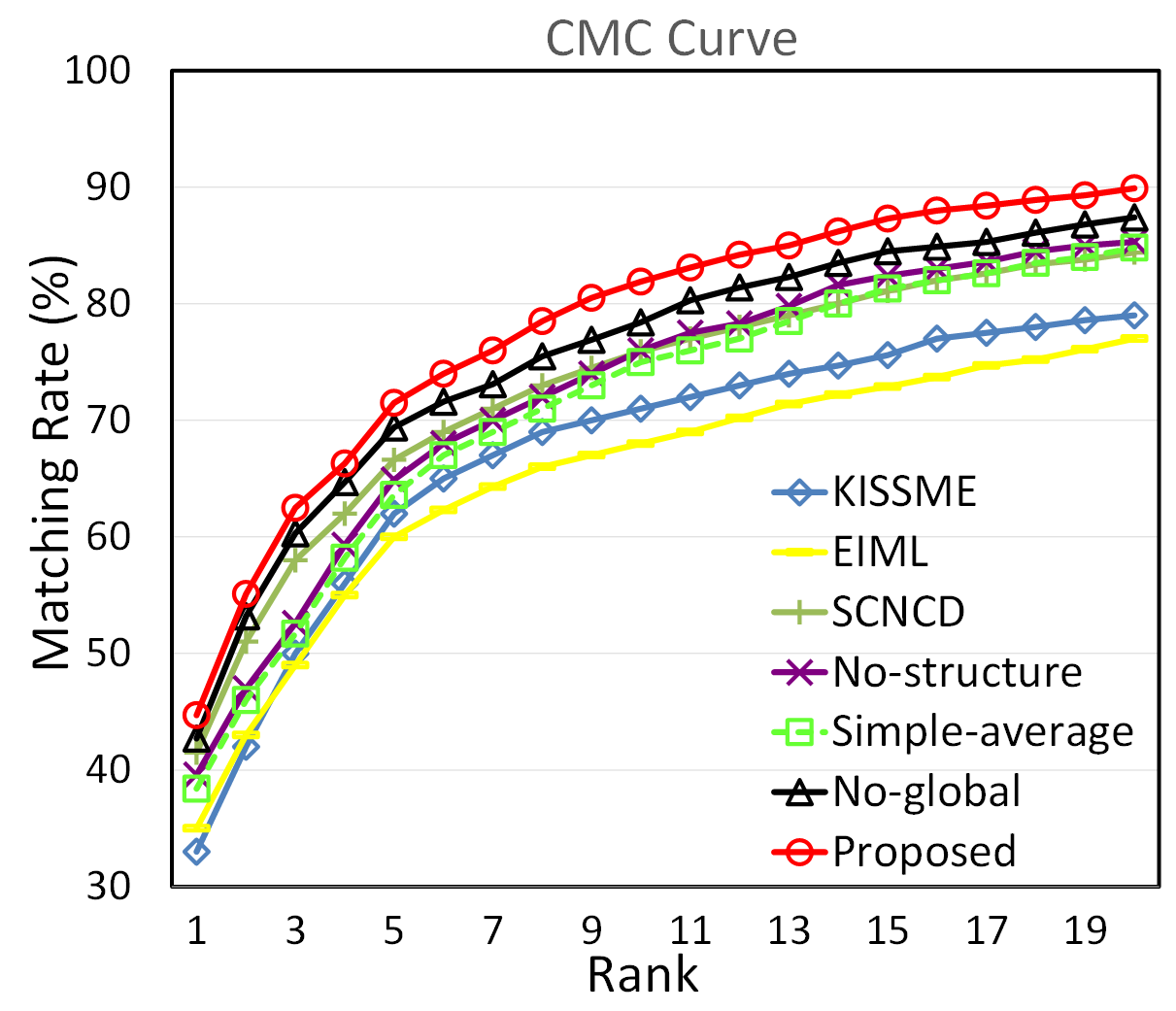}
    \label{fig:cmc curve b}}
    \hspace{3mm}
  \subfloat[the 3DPeS dataset]{\includegraphics[width=5.3cm,height=4.3cm]{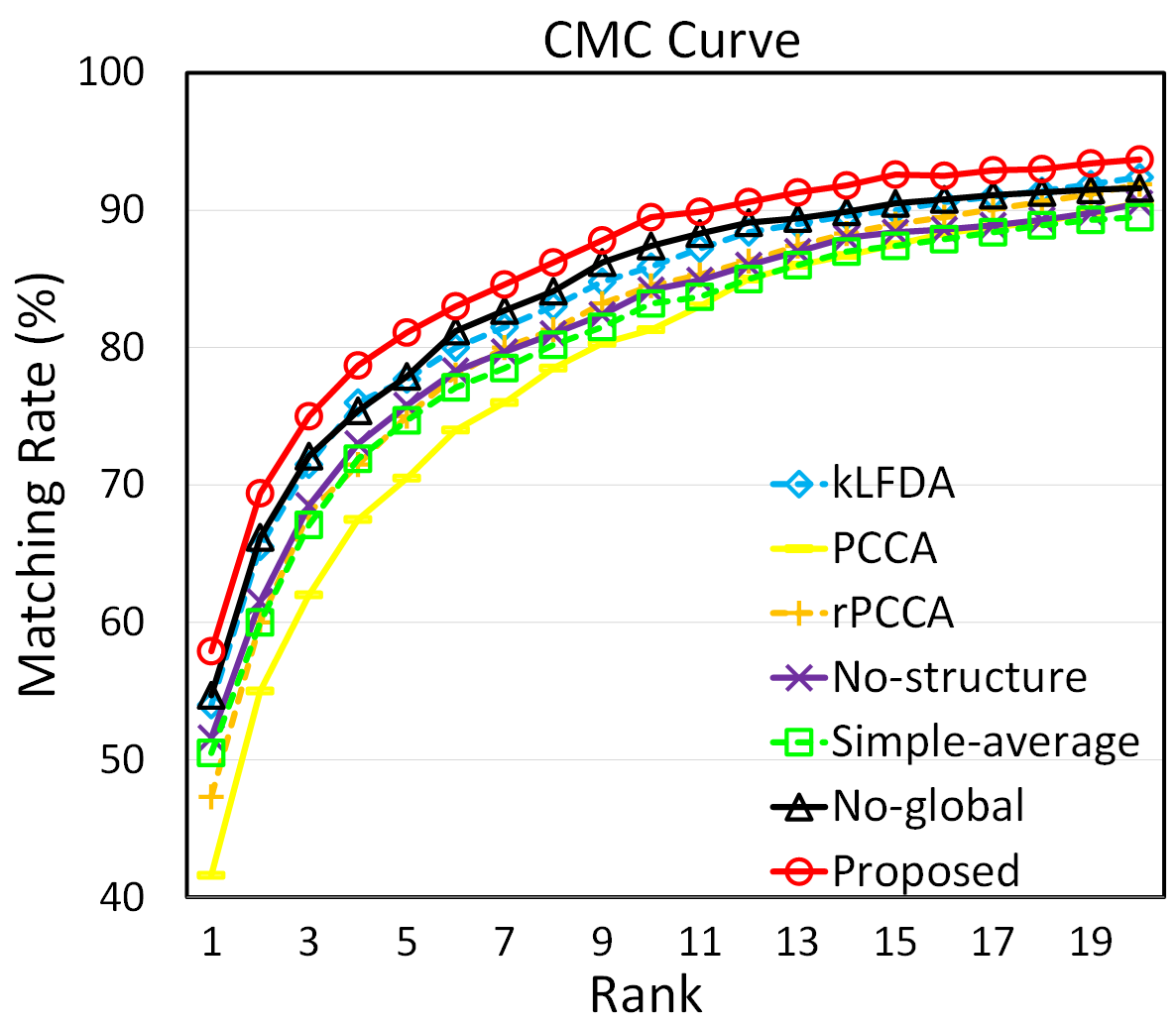}
    \label{fig:cmc curve c}}
  \caption{CMC curves for different methods.}
    \label{fig:cmc curve}
\end{figure*}

We compare results of four methods: (1) Not applying correspondence structure and directly using the appearance similarity between co-located patches for person Re-ID (\emph{No-structure}); (2) Simply averaging the binary mapping structures for different probe images as the correspondence structure and utilizing it for Re-ID (\emph{Simple-average}); (3) Using the correspondence structure learned by our approach, but do not include global constraint when performing Re-ID (\emph{No-global}); (4) Our approach (\emph{Proposed}).

We also compare our results with state-of-the-art methods on different datasets: kLFDA~\cite{kernel-based}, eSDC-ocsvm~\cite{6}, KISSME~\cite{prid}, Salience~\cite{8}, svmml~\cite{svmml}, RankBoost~\cite{rankboost} and LF~\cite{lf} on the VIPeR dataset; KISSME~\cite{prid}, EIML~\cite{eiml}, SCNCD~\cite{SalientColor}, SCNCDFinal~\cite{SalientColor} on the PRID 450S dataset; kLFDA~\cite{kernel-based}, rPCCA~\cite{kernel-based}, PCCA~\cite{PCCA} on the 3DPeS dataset; and eSDC-knn~\cite{6} on the Road dataset.

Tables~\ref{tab:cmcTable1}--\ref{tab:cmcTable4} and Fig.~\ref{fig:cmc curve} show the CMC results of different methods. From the CMC results, we can see that: (1) Our approach has better Re-ID performances than the state-of-the-art methods. This demonstrates the effectiveness of our approach. (2) Our approach has obviously improved results than the no-structure method. This indicates that proper correspondence structures can effectively improve Re-ID performances by reducing patch-wise misalignments. (3) The simple-average method has similar performance to the no-structure method. This implies that unsuitably selected correspondence structures cannot improve Re-ID performance. (4) The no-global method also has good Re-ID performance. This further demonstrates the effectiveness of the correspondence structure learned by our approach. Meanwhile, our approach also has superior performance than the no-global method. This demonstrates the usefulness of introducing global constraint in the patch matching process.

\begin{table}
\centering
\caption{CMC results on the VIPeR dataset}\label{tab:cmcTable1}
\footnotesize{
\label{table1}
\begin{tabular}{|p{2cm}|c|*{5}{c|}c}
\hline
\textbf{Rank}& 1& 5& 10& 20& 30& 50\\
\hline
kLFDA\cite{kernel-based}& 32.3& 65.8& 79.7& 90.9& -& -\\
KISSME\cite{prid}& 27& -& 70& 83& -& 95\\
Salience\cite{8}& 30.2& 52.3& -& -& -& -\\
svmml\cite{svmml}& 30.1& 63.2& 77.4& 88.1& -& -\\
RankBoost\cite{rankboost}& 23.9& 45.6& 56.2& 68.7& -& -\\
eSDC-ocsvm\cite{6}& 26.7& 50.7& 62.4& 76.4& -& -\\
LF\cite{lf}& 24.2& -& 67.1& -& -& 94.1\\
No-structure& 27.5& 57.0& 73.7& 83.9& 87.7& 94.3\\
Simple-average& 28.5& 57.9& 74.1& 84.2& 88.3& 94.6\\
No-global& 30.8& 62.7& 77.5& 88.9& 91.7& 95.6\\
{\bf Proposed}& {\bf 34.8}& {\bf 68.7}& {\bf 82.3}& {\bf 91.8}& {\bf 94.9}& {\bf 96.2}\\
\hline
\end{tabular}}
\end{table}

\begin{table}
\centering
\caption{CMC results on the PRID 450S dataset}\label{tab:cmcTable2}
\footnotesize{
\label{table2}
\begin{tabular}{|p{2cm}|c|*{5}{c|}c|}
\hline
\textbf{Rank}& 1& 5& 10& 20& 30& 50\\
\hline
KISSME\cite{prid}& 33& -& 71& 79& -& 90\\
EIML\cite{eiml}& 35& -& 68& 77& -& 90\\
SCNCD\cite{SalientColor}& 41.5& 66.6& 75.9& 84.4& 88.4& 92.4\\
SCNCDFinal\cite{SalientColor}& 41.6& 68.9& 79.4& 87.8& 91.8& 95.4\\
No-structure& 39.6& 64.9& 76.0& 85.3& 89.3& 93.3\\
Simple-average& 38.2& 63.6& 75.1& 84.9& 88.9& 92.4\\
No-global& 42.7& 69.3& 78.2& 87.4& 91.1& 95.1\\
{\bf Proposed}& {\bf 44.4}& {\bf 71.6}& {\bf 82.2}& {\bf 89.8}& {\bf 93.3}& {\bf 96.0}\\
\hline
\end{tabular}}
\end{table}

\begin{table}
\centering
\caption{CMC results on the 3DPeS dataset}\label{tab:cmcTable3}
\footnotesize{
\begin{tabular}{|p{2cm}|c|*{5}{c|}c|}
\hline
\textbf{Rank}& 1& 5& 10& 15& 20& 30\\
\hline
kLFDA\cite{kernel-based}& 54.0& 77.7& 85.9& -& 92.4& -\\
rPCCA\cite{kernel-based}& 47.3& 75.0& 84.5& -& 91.9& -\\
PCCA\cite{PCCA}& 41.6& 70.5& 81.3& -& 90.4& -\\
No-structure& 51.6& 75.8& 84.2& 88.4& 90.5& 92.6\\
Simple-average& 50.5& 74.7& 83.2& 87.4& 89.5& 92.6\\
No-global& 54.7& 77.9& 87.4& 90.5& 91.6& 93.7\\
{\bf Proposed}& {\bf 57.9}& {\bf 81.1}& {\bf 89.5}& {\bf 92.6}& {\bf 93.7}& {\bf 94.7}\\
\hline
\end{tabular}}
\end{table}

\begin{table}
\centering
\caption{CMC results on the Road dataset}\label{tab:cmcTable4}
\footnotesize{
\label{table3}
\begin{tabular}{|p{2cm}|c|*{5}{c|}c|}
\hline
\textbf{Rank}& 1& 5& 10& 15& 20& 30\\
\hline
eSDC-knn\cite{6}& 52.4& 74.5& 83.7& 88.0& 89.9& 91.8\\
No-structure& 50.5& 80.3& 87.0& 91.3& 94.2& 95.7\\
Simple-average& 49.0& 81.7& 90.4& 92.8& 95.7& 96.2\\
No-global& 58.2& 85.6& 94.2& 97.1& 98.1& 98.6\\
{\bf Proposed}& {\bf 61.5}& {\bf 91.8}& {\bf 95.2}& {\bf 98.1}& {\bf 98.6}& {\bf 99.0}\\
\hline
\end{tabular}}
\end{table}

\section{Conclusion\label{section:conclusion}}

In this paper, we propose a novel framework for addressing the problem of cross-view spatial misalignments in person Re-ID. Our framework consists of two key ingredients: 1) introducing the idea of correspondence structure and learning this structure via a novel boosting method to adapt to arbitrary camera configurations; 2) a constrained global matching step to control the patch-wise misalignments between images due to local appearance ambiguity. Extensive experimental results on benchmark show that our approach achieves the state-of-the-art performance.

Under this framework, our future work is devoted to explore new variants of the two components, such as: 1) designing other correspondence structure learning methods that allow for multiple structure candidates to enhance its flexibility; 2) devising and incorporating edge-to-edge similarity metrics for solving the constrained global matching problem as graph matching \cite{RRWM,HYPR}, which has been proven more effective in many computer vision applications.

{\small
\bibliographystyle{ieee}
\bibliography{egbib}
}
\end{document}